%% file: main.tex
\documentclass[conference,letterpaper]{IEEEtran}
\IEEEoverridecommandlockouts
\usepackage{cite}
\usepackage{amsmath,amssymb,amsfonts}
\usepackage{algorithm}
\usepackage{algorithmic}
\usepackage{graphicx}
\usepackage{dblfloatfix}
\usepackage{float}
\usepackage{multirow}
\usepackage{placeins}
\usepackage{textcomp}
\usepackage{xcolor}
\usepackage{adjustbox}
\definecolor{neongreen}{RGB}{57,255,20}
\usepackage{xurl}
\usepackage[colorlinks=true,linkcolor=neongreen,citecolor=neongreen,urlcolor=neongreen]{hyperref}
\usepackage{etoolbox}




\makeatletter
\long\def\@makecaption#1#2{%
\ifx\@captype\@IEEEtablestring%
\footnotesize\bgroup\par\centering\@IEEEtabletopskipstrut{\normalfont\footnotesize #1}\\{\normalfont\footnotesize #2}\par\addvspace{0.5\baselineskip}\egroup%
\@IEEEtablecaptionsepspace
\else
\@IEEEfigurecaptionsepspace
\setbox\@tempboxa\hbox{\normalfont\footnotesize {#1.}\nobreakspace\nobreakspace #2}%
\ifdim \wd\@tempboxa >\hsize%
\setbox\@tempboxa\hbox{\normalfont\footnotesize {#1.}\nobreakspace\nobreakspace}%
\parbox[t]{\hsize}{\normalfont\footnotesize\noindent\unhbox\@tempboxa#2}%
\else%
\ifCLASSOPTIONconference \hbox to\hsize{\normalfont\footnotesize\hfil\box\@tempboxa\hfil}%
\else \hbox to\hsize{\normalfont\footnotesize\box\@tempboxa\hfil}%
\fi\fi\fi}
\makeatother

\makeatletter
\newcommand\fs@ruledlower{\def\@fs@cfont{\bfseries}\let\@fs@capt\floatc@ruled
  \def\@fs@pre{\vspace{4pt}\hrule height.8pt depth0pt \kern2pt}%
  \def\@fs@post{\kern2pt\hrule\relax}%
  \def\@fs@mid{\kern2pt\hrule\kern2pt}%
  \let\@fs@iftopcapt\iftrue}
\makeatother
\floatstyle{ruledlower}
\restylefloat{algorithm}

\newcommand{\figref}[1]{Fig.~\ref{#1}}
\newcommand{\tabref}[1]{Table~\ref{#1}}
\newcommand{\eqnref}[1]{Eq.~\eqref{#1}}
\newcommand{\sref}[1]{Sec.~\ref{#1}}

\usepackage{tikz}
\usetikzlibrary{shapes.geometric}

\makeatletter
\newcommand{\resetparastyle}{%
  \par
    \normalfont\normalsize\normalcolor
      \leftskip=0pt
        \rightskip=0pt
          \@rightskip=0pt
            \parfillskip=0pt plus 1fil
              \parindent=1em
                \parskip=0pt
                  \spaceskip=0pt
                    \xspaceskip=0pt
                    }
                    \makeatother

                    \AtBeginEnvironment{thebibliography}{%
                      \sloppy\emergencystretch=3em%
                          }

                          \newcommand{\replanning}{replanning}


                            \newcommand{\R}{\mathbb{R}}
                            \newcommand{\E}{\mathbb{E}}
                            \newcommand{\calN}{\mathcal{N}}
                            \newcommand{\bx}{\mathbf{x}}
                            \newcommand{\bu}{\mathbf{u}}
                            \newcommand{\bX}{\mathbf{X}}
                            \newcommand{\bU}{\mathbf{U}}
                            \newcommand{\bc}{\mathbf{c}}
                            
                            \newcommand{\by}{\mathbf{y}}
                            \newcommand{\bmu}{\boldsymbol{\mu}}
                            \newcommand{\bsigma}{\boldsymbol{\sigma}}
                            \newcommand{\bepsilon}{\boldsymbol{\epsilon}}
                            \newcommand{\bsu}{\boldsymbol{\sigma}_u}
                            
                            \newcommand{\bp}{\mathbf{p}}

                            \newcommand{\bV}{\mathbf{V}}
                            \newcommand{\mCollIDM}{11.70}
                            \newcommand{\mOffIDM}{9.10}
                            \newcommand{\mRealIDM}{0.80}
                            \newcommand{\mMinADEIDM}{1.82}

                            \newcommand{\mCollSIMPLAR}{7.30}
                            \newcommand{\mOffSIMPLAR}{6.13}
                            \newcommand{\mRealSIMPLAR}{0.62}
                            \newcommand{\mMinADESIMPLAR}{1.62}
                            \newcommand{\mCollOurs}{4.83 $\pm$ 0.15}
                            \newcommand{\mOffOurs}{3.27 $\pm$ 0.02}
                            \newcommand{\mRealOurs}{0.46 $\pm$ 0.11}
                            \newcommand{\mMinADEOurs}{1.68 $\pm$ 0.013}
                            \newcommand{\mCollOursNoPrior}{5.21 $\pm$ 0.30}
                            \newcommand{\mOffOursNoPrior}{4.0 $\pm$ 0.09}
                            \newcommand{\mRealOursNoPrior}{0.50 $\pm$ 0.01}
                            \newcommand{\mMinADEOursNoPrior}{1.83 $\pm$ 0.004}

                            \newcommand{\rtOursSfive}{91.40 $\pm$ 5.86}
                            \newcommand{\rtOursSten}{231.28 $\pm$ 3.84}
                            \newcommand{\rtOursStwentyfive}{411.15 $\pm$ 2.46}
                            \newcommand{\rtOursGuidedSfive}{128.74 $\pm$ 6.31}
                            \newcommand{\rtOursGuidedSten}{296.42 $\pm$ 5.27}
                            \newcommand{\rtOursGuidedStwentyfive}{525.63 $\pm$ 7.12}
\newcommand{\rtNoPriorSfive}{95.20 $\pm$ 4.10}
\newcommand{\rtNoPriorSten}{238.40 $\pm$ 5.20}
\newcommand{\rtNoPriorStwentyfive}{419.30 $\pm$ 6.40}
                            \newcommand{\mCollOursSfive}{4.83}
                            \newcommand{\mOffOursSfive}{3.27}
                            \newcommand{\mRealOursSfive}{0.46}
                            \newcommand{\mMinADEOursSfive}{1.68}
                            \newcommand{\mCollOursSten}{5.07}
                            \newcommand{\mOffOursSten}{3.52}
                            \newcommand{\mRealOursSten}{0.47}
                            \newcommand{\mMinADEOursSten}{1.73}
                            \newcommand{\mCollOursStwentyfive}{5.61}
                            \newcommand{\mOffOursStwentyfive}{3.94}
                            \newcommand{\mRealOursStwentyfive}{0.50}
                            \newcommand{\mMinADEOursStwentyfive}{1.81}
\newcommand{\mCollNoPriorSfive}{5.21}
\newcommand{\mOffNoPriorSfive}{4.00}
\newcommand{\mRealNoPriorSfive}{0.50}
\newcommand{\mMinADENoPriorSfive}{1.83}
\newcommand{\mCollNoPriorSten}{5.10}
\newcommand{\mOffNoPriorSten}{3.80}
\newcommand{\mRealNoPriorSten}{0.49}
\newcommand{\mMinADENoPriorSten}{1.80}
\newcommand{\mCollNoPriorStwentyfive}{5.02}
\newcommand{\mOffNoPriorStwentyfive}{3.70}
\newcommand{\mRealNoPriorStwentyfive}{0.48}
\newcommand{\mMinADENoPriorStwentyfive}{1.74}
                            \newcommand{\mCollOursGuidedSfive}{4.02}
                            \newcommand{\mOffOursGuidedSfive}{2.46}
                            \newcommand{\mRealOursGuidedSfive}{0.51}
                            \newcommand{\mMinADEOursGuidedSfive}{1.71}
                            \newcommand{\mCollOursGuidedSten}{4.18}
                            \newcommand{\mOffOursGuidedSten}{2.63}
                            \newcommand{\mRealOursGuidedSten}{0.57}
                            \newcommand{\mMinADEOursGuidedSten}{1.74}
                            \newcommand{\mCollOursGuidedStwentyfive}{4.69}
                            \newcommand{\mOffOursGuidedStwentyfive}{3.01}
                            \newcommand{\mRealOursGuidedStwentyfive}{0.69}
                            \newcommand{\mMinADEOursGuidedStwentyfive}{1.80}

                            \title{\LARGE \bf Proposal-Conditioned Latent Diffusion for Closed-Loop Traffic Scenario Generation}

                            \author{
                              \parbox{\textwidth}{%
                                \centering
                                Shubham Vaijanath Phoolari$^{1,3*}$,
                                Aleyna Kara$^{2}$,
                                Christoph Lauer$^{1}$,
                                Steven Peters$^{3}$%
                              }%
                              \thanks{$^{1}$Authors are with CARIAD SE, Germany.}%
                              \thanks{$^{2}$Author is with the Technical University of Munich, Germany.}%
                              \thanks{$^{3}$Author is with the Institute of Automotive Engineering (FZD), TU Darmstadt, Germany.}%
                            }

\begin{document}

                            \maketitle
                            \thispagestyle{empty}
                            \pagestyle{empty}

                            \begin{abstract}
Closed-loop traffic simulation remains challenging because it must generate interactive multi-agent behaviors that are scene-consistent and controllable during rollout. Prior diffusion-based approaches yield strong realism but their computational cost can hinder deployment in time-constrained replanning loops for AV planning and simulation. We present a diffusion-based scenario generation framework that conditions on instance-centric scene context and multi-modal proposal priors with optional test-time guidance for shaping safety-critical behaviors. A compact action-latent parameterization and proposal-based initialization improve sampling efficiency and reduce per-step runtime without retraining. Experiments on the Waymo Open Motion Dataset indicate a favorable balance between realism, safety and controllability across diverse interactive scenarios and show that guidance enables systematic trade-offs across objectives.
                            \end{abstract}

                            \section{Introduction}
                            \label{sec:introduction}
Validating autonomous vehicles, AVs, demands systematic exposure to safety-critical events that, by definition, occur rarely in naturalistic driving data~\cite{wachenfeld2016release}. On-road testing is expensive and slow while simulated testing provides an inexpensive alternative as long as the simulated traffic is sufficiently realistic and includes a diverse set of meaningful edge cases.

                            Prior research has contributed to the development of multi-agent scenario generation methodologies along two complementary directions. Realism-focused methods learn data-driven priors from large-scale datasets and generate highly realistic multi-agent behavior~\cite{suo2021trafficsimlearningsimulaterealistic,feng2023trafficgenlearninggeneratediverse}. However, these methods provide little control over the safety-critical aspects of the generated scenarios. Conversely, adversarial methods focus on generating scenarios that include collisions or near-miss events~\cite{rempe2022generatingusefulaccidentpronedriving,hanselmann2022kinggeneratingsafetycriticaldriving} and may prioritize collision events at the expense of plausibility, such as agents violating road boundaries or kinematic constraints. Bridging this dichotomy is essential. Effective verification and validation requires scenarios that are simultaneously challenging and plausible to yield actionable safety insights.

                            Controllable multi-agent motion generation using diffusion models presents a potential solution to the challenge of providing both controllability and plausibility. The iterative nature of diffusion model-based denoising facilitates the use of test-time guidance and objective composition without the need for retraining~\cite{jiang2023motiondiffusercontrollablemultiagentmotion,huang2024vbd}. Nevertheless, reverse sampling typically involves many sequential steps which limits throughput in closed-loop environments where \replanning\ must occur frequently and can make runtime efficiency a bottleneck for AV planning and simulation. Furthermore, joint guidance across multiple agents increases the computational complexity associated with the diffusion process and can lead to kinematically inconsistent trajectories. Starting the reverse process from a data-informed Gaussian prior can reduce the number of steps while preserving closed-loop fidelity~\cite{wang2024opttrajdiff}.

                            % Force start of right column on page 1.

                            % --- Fig. 1 at top of right column ---
                            \begin{figure}[!t]
                              \centering
                                \setlength{\abovecaptionskip}{2pt}
                                  \setlength{\belowcaptionskip}{0pt}
                                    % Crop white margins in the PNG so visible content aligns better with caption/text
                                      \includegraphics[width=\columnwidth,trim=10 8 10 10,clip]{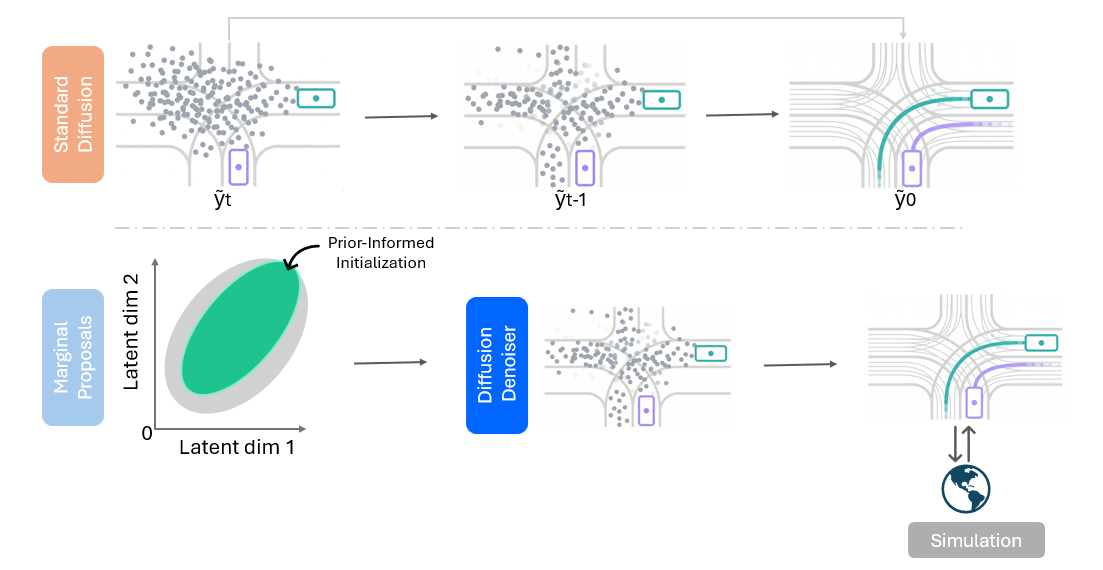}
                                        \caption{Our approach is proposal-informed diffusion for closed-loop traffic simulation, initializing sampling from instance-centric marginal priors.}
                                          \label{fig:overview_firstpage}
                                          \end{figure}

                                          % Unbreakable contributions block (heading + bullets stay together)
                                          \noindent\begin{minipage}[t]{\columnwidth}
                                          \vspace{0pt}
                                          \textbf{Our key contributions:}
                                          \begin{itemize}
                                            \setlength{\itemsep}{1pt}
                                              \setlength{\parskip}{0pt}
                                                \setlength{\parsep}{0pt}
                                                  \setlength{\topsep}{2pt}
                                                    \item We propose a proposal-conditioned joint diffusion policy for closed-loop simulation that conditions on instance-centric scene context and per-agent marginal proposals, explicitly modeling joint interaction rather than composing independent futures.
                                                    \item We map proposal trajectories to a compact action latent via PCA and use proposal statistics to construct a shifted Gaussian start distribution, enabling few-step reverse diffusion at test time without retraining.
                                                        \item We apply differentiable map, collision, and game-theoretic objectives as latent-space guidance to trade realism vs stress-testing in closed-loop rollouts.
                                                        \end{itemize}
                                                        \end{minipage}

                                                        \input{related_work}

                                                                \begin{figure*}[!t]
                                                                  \centering
                                                                    \includegraphics[width=\textwidth]{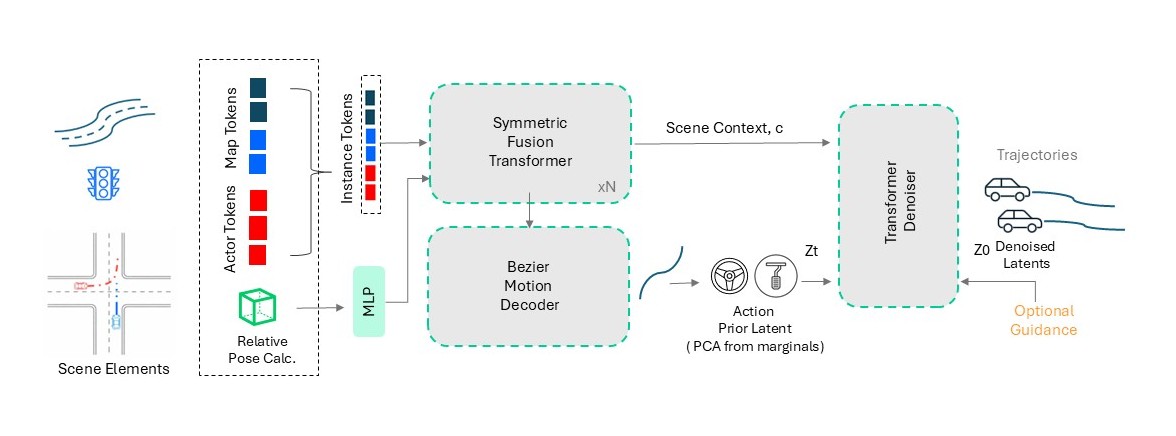}
                                                                      \caption{Our architecture illustration. An instance-centric symmetric scene encoder maps agent history and map context into a scene context $c$ and marginal futures that are converted to an action-prior latent via PCA to condition a joint Transformer denoiser for multi-agent trajectory generation. We optionally apply inference-time guidance to refine sampled plans for behavior control and safety-critical scenarios.}
                                                                        \label{fig:simpl_arch}
                                                                        \end{figure*}

                                                                \section{Methodology}
                                                                \label{sec:method}
                                                                We first summarize the architecture in \figref{fig:simpl_arch}, then detail the formulation and inference procedure.

                                                                % --- Tighter, cleaner displayed-equation spacing for IEEE ---
                                                                \setlength{\abovedisplayskip}{4pt}
                                                                \setlength{\belowdisplayskip}{4pt}
                                                                \setlength{\abovedisplayshortskip}{2pt}
                                                                \setlength{\belowdisplayshortskip}{2pt}
                                                                % Extra vertical separation between consecutive display equations
                                                                \setlength{\jot}{3pt}

                                                                \subsection{Problem formulation}
                                                                We consider a simulated traffic scenario with $N$ agents. Agent $i$ at physical time $t$ has state
                                                                \begin{equation}
                                                                \bx_t^i = [x_t^i,\ y_t^i,\ \psi_t^i,\ v_t^i]^\top \in \R^{4},
                                                                \end{equation}
                                                                where $(x_t^i,y_t^i)$ is planar position, $\psi_t^i$ is heading, and $v_t^i$ is speed. The action is
                                                                \begin{equation}
                                                                \bu_t^i = [a_t^i,\ \dot{\psi}_t^i]^\top \in \R^{2},
                                                                \end{equation}
                                                                where $a_t^i$ is longitudinal acceleration and $\dot{\psi}_t^i$ is yaw rate. We distinguish an ego agent that follows a fixed known policy and a set of reactive agents. At each replanning time $\tau$, the simulator provides the current closed loop state $\bX_{\mathrm{init}}(\tau)$ together with observation history $H_\tau$ and map context $M_\tau$. Our goal is to learn a reactive policy $g$ that maps the current state, history, and map context to a joint action plan for the reactive agents over a horizon of $T_u$ steps. The joint plan is
                                                                \begin{equation}
                                                                \bU_\tau = \{\bu_{\tau+t}^i\}_{i\in\mathcal{R},\ t\in\{1,\dots,T_u\}} \in \R^{|\mathcal{R}|\times T_u\times 2},
                                                                \end{equation}
                                                                where $\mathcal{R}$ denotes the set of reactive agents. Equivalently, $g:(\bX_{\mathrm{init}}(\tau),H_\tau,M_\tau)\mapsto \R^{|\mathcal{R}|\times T_u\times 2}$. The policy maps the current closed loop situation to a plan
                                                                \begin{equation}
                                                                \bU_\tau = g\!\left(\bX_{\mathrm{init}}(\tau),\, H_\tau,\, M_\tau\right),
                                                                \end{equation}
                                                                and the output of $g$ is a finite horizon action sequence. The generated actions are rolled out by a differentiable kinematic model
                                                                \begin{equation}
                                                                \hat{\bX}_\tau = f\!\left(\bX_{\mathrm{init}}(\tau),\, \bU_\tau\right),
                                                                \label{eq:rollout}
                                                                \end{equation}
                                                                with sampling period $\Delta t$. In closed loop simulation, the ego agent executes its fixed policy, the reactive agents execute the first part of $\bU_\tau$, and replanning repeats at the next replanning time.

                                                                \subsection{Instance-Centric Scene Encoding}
                                                                \label{subsec:simpl_scene}
                                                                We use an instance-centric scene encoding that represents agents and map elements in a shared token space and produces both scene context features and per-agent marginal trajectory proposals. We provide a brief explanation of this method here and refer to~\cite{zhang2024simpl} for a more detailed introduction. Each scene is decomposed into a set of agent instances and a set of map polyline instances. Each instance is expressed in its own local frame. For an agent instance, the local frame origin is the last observed pose at replanning time $\tau$ and the local $x$ axis is aligned with the agent heading at that time. The agent input features include a fixed length history of motion and kinematics in this local frame. For a map polyline instance, the local frame origin is the polyline centroid. The local $x$ axis is aligned with the polyline tangent direction defined by the polyline geometry. In our implementation, the tangent is the unit direction induced by the ordered polyline points from first to last after consistent ordering. Traffic light state is represented as a categorical attribute of lane elements and is embedded into the polyline features.

                                                                The encoder produces agent tokens $Z^{(\mathrm{a})}=\{z_i^{(\mathrm{a})}\}_{i=1}^{N_a}$ and map tokens $Z^{(\mathrm{m})}=\{z_j^{(\mathrm{m})}\}_{j=1}^{N_m}$, concatenated as $Z\in\R^{N_z\times D}$ with $N_z=N_a+N_m$, where each of the $N_z$ rows is a $D$ dimensional instance embedding. Token interactions are parameterized by a relative positional encoding. Let $c_i\in\R^2$ denote the instance center in a common scene frame and $q_i\in\R^2$ the unit direction associated with instance $i$ (agent heading or lane tangent). We define the displacement $p_{i\to j}=c_j-c_i$ and the relation descriptor
                                                                \begin{equation}
                                                                r_{ij}=\big[\|p_{i\to j}\|_2,\ \angle(q_i,p_{i\to j}),\ \angle(q_i,q_j)\big],
                                                                \end{equation}
                                                                where $\angle(\cdot,\cdot)$ returns a signed wrapped angle in $(-\pi,\pi]$. An MLP maps $r_{ij}$ to an edge embedding $e_{ij}\in\R^D$. A stack of Symmetric Fusion Transformer layers, abbreviated SFT, refines $Z$ using attention conditioned on these relational embeddings. This yields a scene context representation $\hat{\bc}_\tau$. The encoder also outputs $K$ marginal future trajectory proposals per agent with associated scores.

                                                                \subsection{Marginal priors and PCA latent construction}
                                                                Given the $K$ proposal futures and their probabilities, each proposal trajectory is mapped to an action sequence using an inverse dynamics routine consistent with the rollout model. Each action is $\bu=[a,\dot{\psi}]^\top$, resulting in a flattened interleaved vector
                                                                \begin{equation}
                                                                \bu^i = [a_1^i,\dot{\psi}_1^i,\dots,a_{T_u}^i,\dot{\psi}_{T_u}^i]\in\R^{D_u},
                                                                \end{equation}
                                                                where $D_u=2T_u$ is the action dimension per agent over the horizon. Action sequences are compressed into a low-dimensional latent space using principal component analysis (PCA). We compute global affine statistics $(\bmu_u,\bsu)$ over all action sequences in the training set (across agents and proposals) and fix them for normalization. With these statistics and PCA components $\bV\in\R^{D_u\times d}$,
                                                                \begin{equation}
                                                                \by^i = \left((\bu^i-\bmu_u)\oslash\bsu\right)\bV \in\R^{d},\qquad d\ll D_u,
                                                                \end{equation}
                                                                where $\oslash$ denotes elementwise division. This process yields (i) a latent target $\by_0$ from the ground-truth action sequence and (ii) for each agent $i$, a set of latent proposal modes $\{\by^{i,(k)}\}_{k=1}^K$. Across all agents, the marginal prior set is $\{ \by^{i,(k)} \mid i=1..N,\, k=1..K \}$.

                                                                \subsection{Latent diffusion denoiser}
                                                                \label{subsec:latent_denoiser}
                                                                A diffusion model generates a clean latent plan by learning to reverse a gradual noising process. The clean joint latent plan is $\by_0 \in \R^{N\times d}$. For each diffusion step $s$, we define a noisy version $\tilde{\by}_s$ obtained by applying a variance preserving forward process to $\by_0$.

                                                                \smallskip
                                                                \noindent\textbf{b) Forward process and training objective.}
                                                                We use a variance preserving diffusion process defined by a noise schedule $\{\beta_s\}_{s=1}^{S}$. We set
                                                                \begin{equation}
                                                                \alpha_s = 1-\beta_s,\qquad \bar{\alpha}_s = \prod_{j=1}^{s}\alpha_j.
                                                                \label{eq:diff_schedule}
                                                                \end{equation}
                                                                The quantity $\bar{\alpha}_s$ determines how much of the clean signal remains at step $s$. Small $s$ yields $\bar{\alpha}_s$ close to one and $\tilde{\by}_s$ close to $\by_0$. Large $s$ yields $\bar{\alpha}_s$ close to zero and $\tilde{\by}_s$ close to noise.

                                                                We initialize the noise scale using the marginal proposal set produced by the scene encoder. For agent $i$ we have proposal latents $\{\by_{i,(k)}\}_{k=1}^{K}\subset \R^{d}$. We compute per dimension proposal statistics
                                                                \begin{equation}
                                                                \begin{aligned}
                                                                \bmu_i &= \frac{1}{K}\sum_{k=1}^{K}\by_{i,(k)},\\
                                                                \bsigma^{\text{train}}_{y,i}
                                                                &= \mathrm{Std}_{k}\!\left(\by_{i,(k)}\right) + \epsilon_\sigma,
                                                                \qquad \epsilon_\sigma>0.
                                                                \end{aligned}
                                                                \label{eq:proposal_stats}
                                                                \end{equation}
                                                                Stacking across agents gives
                                                                \begin{equation}
                                                                \bmu \in \R^{N\times d},\qquad
                                                                \bsigma^{\text{train}}_{y} \in \R^{N\times d}.
                                                                \label{eq:stacked_stats}
                                                                \end{equation}
                                                                The forward noising process is
                                                                \begin{equation}
                                                                \tilde{\by}_s = \sqrt{\bar{\alpha}_s}\by_0 + \sqrt{1-\bar{\alpha}_s}\,\big(\bsigma^{\text{train}}_{y}\odot\bepsilon\big),
                                                                \label{eq:train_summary_forward}
                                                                \end{equation}
                                                                where $\bepsilon \sim \calN(\mathbf{0},\mathbf{I})$ and $\odot$ denotes elementwise multiplication. This equation defines the noised latent under the schedule $\{\bar{\alpha}_s\}$; it is obtained by injecting step-dependent noise according to $\bar{\alpha}_s$.

                                                                The denoiser $D_\theta(\tilde{\by}_s, s, \hat{\bc}_\tau,\mathcal{P}_\tau)$ predicts the scaled noise term $\bsigma^{\text{train}}_{y}\odot\bepsilon$ from the noisy input $\tilde{\by}_s$ and the conditioning. We train it with the standard DDPM noise prediction loss~\cite{ho2020ddpm}
                                                                \begin{equation}
                                                                \mathcal{L}_{\mathrm{diff}}
                                                                = \E_{(\mathrm{scene}),\,s,\,\bepsilon}\!\Big[\big\|\bsigma^{\text{train}}_{y}\odot\bepsilon - D_\theta(\tilde{\by}_s, s, \hat{\bc}_\tau,\mathcal{P}_\tau)\big\|_2^2\Big].
                                                                \label{eq:train_x0hat_onestep}
                                                                \end{equation}
                                                                At inference we optionally use DDIM to reduce the number of denoising steps while keeping the same denoiser~\cite{song2020ddim}.

                                                                \smallskip
                                                                \noindent\textbf{Inverse PCA map used for rollout.}
                                                                Given a noisy latent, we form the one step estimate of the clean latent
                                                                \begin{equation}
                                                                \hat{\by}_{0\mid s}
                                                                =
                                                                \frac{
                                                                \tilde{\by}_s
                                                                - \sqrt{1-\bar{\alpha}_s}\;
                                                                D_\theta\!\left(\tilde{\by}_s, s, \hat{\bc}_\tau,\mathcal{P}_\tau\right)
                                                                }{\sqrt{\bar{\alpha}_s}},
                                                                \label{eq:inference_x0hat_onestep}
                                                                \end{equation}
                                                                and map it back to action sequences using the fixed PCA inverse and fixed affine statistics computed once from the training set. For each agent $i$,
                                                                \begin{equation}
                                                                \hat{\bu}^i = (\hat{\by}^i \bV^\top)\odot\bsigma_u + \bmu_u,
                                                                \label{eq:train_summary_decode}
                                                                \end{equation}
                                                                where $\hat{\by}^i \in \R^{d}$, $\bV\in\R^{D_u\times d}$, and $\hat{\bu}^i\in\R^{D_u}$. The vector is then reshaped into $\{\hat{\bu}_t^i\}_{t=1}^{T_u}$ with $\hat{\bu}_t^i=[\hat{a}_t^i,\ \dot{\hat{\psi}}_t^i]^\top$. We then roll out
                                                                \begin{equation}
                                                                \hat{\bX} = f(\bX_{\mathrm{init}},\hat{\bU}).
                                                                \label{eq:train_summary_rollout}
                                                                \end{equation}
                                                                We define a state matching loss on the rollout
                                                                \begin{equation}
                                                                \mathcal{L}_{\mathrm{state}}
                                                                = \E_{(\mathrm{scene}),\,s,\,\bepsilon}\!\left[
                                                                \mathrm{SmoothL1}\!\left(\hat{\bX}^{(x,y,\psi)},\,\bX_{\mathrm{gt}}^{(x,y,\psi)}\right)
                                                                \right],
                                                                \label{eq:train_summary_lstate}
                                                                \end{equation}
                                                                and combine it with the diffusion loss as, 
                                                                \vspace{2pt}
                                                                \begin{equation}
                                                                \mathcal{L}
                                                                 = \lambda_{\mathrm{diff}}\,\mathcal{L}_{\mathrm{diff}}
                                                                 {}+ \lambda_{\mathrm{state}}\,\mathcal{L}_{\mathrm{state}},
                                                                 \label{eq:train_summary_total}
                                                                 \end{equation}
                                                                 \noindent where $\lambda_{\mathrm{state}}=0.5$ and $\lambda_{\mathrm{diff}}=1$.
                                                                 \par\smallskip
                                                                 \noindent\textbf{c) Shifted Gaussian initialization from proposal statistics.}
                                                                 Motivated by OptTrajDiff's proposal-informed Gaussian initialization~\cite{wang2024opttrajdiff}, we warm-start the reverse process from a Gaussian whose mean and diagonal scale are given by \eqnref{eq:proposal_stats}. At reverse step $S$, we sample
                                                                 \begin{equation}
                                                                 \tilde{\by}_S
                                                                 = \sqrt{\bar{\alpha}_S}\,\bmu
                                                                 + \sqrt{1-\bar{\alpha}_S}\,\big(\bsigma^{\text{train}}_{y}\odot\bepsilon_S\big),\qquad
                                                                 \bepsilon_S\sim \calN(\mathbf{0},\mathbf{I}).
                                                                 \label{eq:prior_init_sample}
                                                                 \end{equation}

                                                                 \subsection{Guidance at inference}
                                                                 \label{subsec:guidance}
                                                                 We optionally refine sampled plans using gradient-based guidance in latent space at inference time.

                                                                 \noindent Here, $\hat{\by}_{0\mid s}$ is the one-step analytic estimate of the clean latent implied by the denoiser at the sampled diffusion step $s$.

                                                                 \smallskip
                                                                 \noindent\textbf{Objective-based guidance.}
                                                                 We define a cost $J(\hat{\bX})$ on the rolled-out trajectories and refine the sampled plan to reduce it. In our experiments,
                                                                 \begin{equation}
                                                                 J(\hat{\bX}) = w_{\mathrm{map}} J_{\mathrm{map}}(\hat{\bX}) + w_{\mathrm{coll}} J_{\mathrm{coll}}(\hat{\bX}),
                                                                 \label{eq:guidance_cost}
                                                                 \end{equation}
                                                                where $J_{\mathrm{map}}$ penalizes map violations using an off-road signed-distance proxy and $J_{\mathrm{coll}}$ penalizes near-term overlaps using a smooth distance-based proxy. At inference, we decode the denoiser output to actions, roll out $\hat{\bX}$, and apply a gradient step that updates the latent plan:
                                                                 \begin{equation}
                                                                 \hat{\by}_0 \leftarrow \hat{\by}_0 - \eta \nabla_{\hat{\by}_0}
                                                                 J\!\left(f\!\left(\bX_{\tau}, \mathrm{decode}(\hat{\by}_0)\right)\right)
                                                                 \label{eq:guidance_update}
                                                                 \end{equation}
                                                                 optionally, only in late reverse steps and without backpropagating through $D_{\theta}$ across diffusion steps.

                                                                 \noindent The guidance  (i) tightens constraint satisfaction in rare corner cases and (ii) performs targeted, counterfactual edits without retraining. The same mechanism can incorporate additional differentiable objectives.

                                                                \smallskip
                                                                \noindent\textbf{Safety-critical game-theoretic guidance.}
                                                                We model safety-critical interactions as a pursuit--evasion game with an IBR-style refinement during guided sampling~\cite{huang2024vbd}. Agents are split into ego-controlled and adversarial sets, with costs $J_{\mathrm{ego}}(\hat{\bX})$ and $J_{\mathrm{adv}}(\hat{\bX})$. From the current latent estimate $\hat{\by}_0$, we alternate a few best-response updates: adversaries increase $J_{\mathrm{ego}}$, ego agents decrease it, and each step decodes and rolls out before computing gradients. This adapts both sides and produces interactive, scene-consistent adversarial scenarios.

                                                                 \begin{algorithm}[t]
                                                                 \setlength{\abovecaptionskip}{0pt}
                                                                 \setlength{\belowcaptionskip}{1pt}
                                                                 \fontsize{7.4}{7.75}\selectfont
                                                                 \caption{Closed-loop simulation with latent diffusion policy}
                                                                 \label{alg:closedloop}
                                                                 \begin{algorithmic}[1]
                                                                 \setlength{\itemsep}{0pt}\setlength{\parsep}{0pt}
                                                                 \STATE \textbf{Input.} initial state $\bX_0$; replanning interval $\Delta r$; horizon $T_u$; ego policy $\pi_{\mathrm{ego}}$; rollout model $f$; SIMPL encoder; denoiser $D_\theta$
                                                                 \FOR{$\tau = 0,\Delta r,2\Delta r,\dots$}
                                                                   \STATE Encode the current scene at $\bX_\tau$ to obtain context $\hat{\bc}_\tau$ and marginal proposals $\{(\by_\tau^{(k)},\pi_\tau^{(k)})\}_{k=1}^{K}$
                                                                   \STATE Sample a joint latent plan $\hat{\by}_{0,\tau}$ by reverse diffusion conditioned on $\hat{\bc}_\tau$ and $\{(\by_\tau^{(k)},\pi_\tau^{(k)})\}_{k=1}^{K}$
                                                                   \STATE Decode $\hat{\by}_{0,\tau}$ to a joint action plan $\hat{\bU}_\tau=\{\hat{\bu}^{i}_{t}\}_{i=1..N,\,t=1..T_u}$
                                                                   \STATE Roll out $\hat{\bX}_\tau=f(\bX_\tau,\hat{\bU}_\tau)$
                                                                   \STATE Execute the first $\Delta r$ steps: ego uses $\pi_{\mathrm{ego}}$, reactive agents use $\hat{\bU}_\tau$
                                                                 \ENDFOR
                                                                 \end{algorithmic}
                                                                 \end{algorithm}

                                                                           \section{Experiments}
                                                                           \label{sec:experiments}

                                                                           \subsection{Dataset}
                                                                           The experiments utilize the Waymo Open Motion Dataset (WOMD)~\cite{ettinger2021womd}, which contains over 500 hours of driving data collected across multiple cities in the United States. Each scenario comprises a 9-second segment, including 1 second of history and an 8-second prediction horizon. We train the diffusion model on a randomly sampled subset of 200k scenarios for efficiency. Closed-loop ablations utilize 500 randomly selected scenarios from the WOMD validation split.

                                                                           \smallskip
                                                                           \noindent\textbf{Diffusion training details.}
                                                                           We use two-stage training: a marginal proposal model first provides SIMPL context and per-agent multi-modal proposals, then a joint latent diffusion model is trained on the resulting proposal-derived priors.
                                                                           The second-stage DDPM uses a variance-preserving process over the joint standardized PCA latent plan with $d=32$, $S=100$ diffusion steps and a linear $\beta$ schedule $\beta_s\in[10^{-4},5\cdot 10^{-2}]$.
                                                                           The Transformer denoiser uses hidden size 128, 8 attention heads, dropout 0.3, $\lambda_{\mathrm{state}}=0.5$ and $\lambda_{\mathrm{diff}}=1$.
                                                                           We use AdamW with learning rate $5e{-4}$, weight decay $10^{-3}$ and bfloat16 mixed precision.

                                                                           \subsection{Baselines and Training Setup}
                                                                           \label{subsec:baselines_train}

                                                                           We select baselines to isolate the main modeling choices: IDM provides a low-compute rule-based reference, SIMPL-AR tests marginal proposal selection without joint diffusion, the unguided joint diffusion baseline isolates guidance, and the context-only ablation removes proposal-informed initialization. All methods use the state/action definitions from \sref{sec:method} and the rollout model $f$ from \eqnref{eq:rollout}. All learned models are trained on a single NVIDIA H100 GPU.

                                                                           \subsubsection{IDM (rule-based baseline)}
                                                                           The IDM baseline is a lightweight deterministic reference. Each non-ego agent follows IDM-style longitudinal control along an assigned lane centerline with heuristic car-following and a default desired speed, without SIMPL predictions or diffusion sampling. This is a simple rule-based reference rather than a calibrated microscopic traffic simulator.

                                                                           \subsubsection{SIMPL-AR (collision-aware selection)}
                                                                           At each replanning time $\tau$, SIMPL predicts $K=6$ multi-modal futures per reactive agent with confidence scores. We convert each proposal to actions via inverse dynamics, roll out with $f$, and select one mode per agent by approximately minimizing
                                                                           \begin{equation}
                                                                           \resizebox{0.98\columnwidth}{!}{$\displaystyle
                                                                           \min_{k_1,\dots,k_N}\sum_{i=1}^{N}-\log\!\left(\max\!\left(p^{i,(k_i)}_{\tau},\epsilon\right)\right)
                                                                           +\lambda_{\mathrm{coll}}\sum_{1\le i<j\le N}\!\mathrm{Coll}\!\left(\tilde{\bX}^{i,(k_i)}_{\tau},\tilde{\bX}^{j,(k_j)}_{\tau}\right)
                                                                           $}
                                                                           \label{eq:simpl_ar_objective}
                                                                           \end{equation}
                                                                           with a lightweight circle-overlap proxy
                                                                           \begin{equation}
                                                                           \resizebox{0.98\columnwidth}{!}{$\displaystyle
                                                                           \mathrm{Coll}\!\left(\tilde{\bX}^{i,(k_i)}_{\tau},\tilde{\bX}^{j,(k_j)}_{\tau}\right)
                                                                           =\sum_{t=1}^{T_{\mathrm{look}}}\mathbb{I}\!\left[\left\|\bp^{i,(k_i)}_{\tau,t}-\bp^{j,(k_j)}_{\tau,t}\right\|_2<r_i+r_j\right]
                                                                           $}
                                                                           \label{eq:collision_proxy}
                                                                           \end{equation}
                                                                           where $p^{i,(k_i)}_{\tau}$ is the confidence score of the selected mode, $\lambda_{\mathrm{coll}}$ weights the collision penalty, and $r_i$ is a circle radius derived from the agent footprint. $T_{\mathrm{look}}$ is the collision lookahead horizon.
                                                                           This SIMPL-AR setup is consistent with a prior WOSAC-style simulator that builds autoregressive closed-loop execution on top of multi-modal motion predictors plus lightweight rollout-time consistency checks~\cite{qian20232ndplacesolution2023}.

                                                                           \subsubsection{Joint diffusion baseline (no guidance)}
                                                                           The stage-2 baseline uses the same architecture and training setup as the main model and is trained for 25 epochs. At inference, it samples a joint latent plan $\by\in\R^{N\times 32}$ conditioned on SIMPL scene context $\hat{\bc}_{\tau}$ and the marginal prior set, then decodes to actions and rolls out with $f$.

                                                                           \subsubsection{Diffusion without prior initialization (context-only)}
                                                                           This ablation removes proposal-informed prior initialization and marginal-prior tokens while still conditioning on the SIMPL scene encoding. Reverse diffusion is initialized from a standard Gaussian rather than the shifted Gaussian in \eqnref{eq:prior_init_sample}. The same stage-2 architecture is trained for 25 epochs. 

\subsection{Closed-loop evaluation}
                                                                           We evaluate with receding-horizon closed-loop simulation in Waymax~\cite{gulino2023waymax}. At each replanning time $\tau$, we sample a joint $8$\,s rollout for all reactive agents (up to $N=64$ closest to SDC), execute the first control step at $10$\,Hz, and then replan from the updated simulator state.

                                                                           \subsection{Metrics}
                                                                           We report standard closed-loop safety and fidelity metrics aligned with prior controllable simulation work: collision rate for interaction feasibility, off-road rate for map consistency, minADE for positional accuracy, and Wasserstein-1 acceleration/jerk histograms for kinematic naturalness. We do not report Waymo Open Sim Agents Challenge metrics, which require the official evaluator and large-scale submission protocol.

                                                                           \begin{itemize}
                                                                             \item \textbf{Off-road.} A binary flag indicating that an agent departs the drivable road surface. We determine this using the agent position relative to oriented road-graph (lane-edge) reference points and report the off-road rate as the fraction of agents that become off-road at any point during rollout, aggregated over all agents and generated scenarios.
                                                                               \item \textbf{Collision.} Collision rate, reported as the fraction of agents that participate in at least one collision during rollout.
                                                                                 \item \textbf{Realism.} Following prior work~\cite{chang2024safesim,zhong2023guided}, we report the average Wasserstein-1 distance between \emph{L1-normalized} histograms of longitudinal acceleration, lateral acceleration and jerk computed from generated trajectories and ground-truth trajectories, aggregated over all agents and timesteps.
                                                                                   \item \textbf{Best-of-$N$ displacement error (minADE).} For each scene, we compute the minimum average $\ell_2$ displacement error to the logged future over $N$ stochastic rollouts, then average over valid agents and scenes.
                                                                                   \end{itemize}

\noindent\textbf{Accelerated sampling and runtime.}
To reduce test-time latency, we use DDIM~\cite{song2020ddim} sampling and vary the number of reverse steps to expose an explicit balance between speed and quality. We report per-\replanning\ runtime together with safety and realism metrics in \tabref{tab:accel_steps}.

% ---- results float block ----
\begin{table*}[!t]
                                                                                   \caption{Closed-loop simulation metrics. For diffusion models we report mean and standard deviation over 5 random seeds. Deterministic baselines are shown without variance.}
                                                                                   \label{tab:metrics}
                                                                                   \centering
                                                                                   \small
                                                                                           \setlength{\tabcolsep}{4pt}
                                                                                   \renewcommand{\arraystretch}{1.0}
                                                                                   \resizebox{0.87\textwidth}{!}{%
\begin{tabular}{lcccc}
                                                                                   \hline
                                                                                   \textbf{Method} & \textbf{Collision rate [\%]} $\downarrow$ & \textbf{Off-road rate [\%]} $\downarrow$ & \textbf{Realism} $\downarrow$ & \textbf{minADE [m]} $\downarrow$\\
                                                                                   \hline
                                                                                   IDM (Rule-based) & \mCollIDM & \mOffIDM & \mRealIDM & \mMinADEIDM\\
                                                                                   SIMPL-AR (collision-aware selection) & \mCollSIMPLAR & \mOffSIMPLAR & \mRealSIMPLAR & \mMinADESIMPLAR\\
                                                                                   Ours (w/o Guidance) & \textbf{\mCollOurs} & \textbf{\mOffOurs} & \textbf{\mRealOurs} & \textbf{\mMinADEOurs}\\
                                                                                   Ours \textsuperscript{\textdagger} (w/o Prior Init) & \mCollOursNoPrior & \mOffOursNoPrior & \mRealOursNoPrior & \mMinADEOursNoPrior\\
                                                                                   \hline
                                                                                   \end{tabular}
}
                                                                                   \par\vspace{0mm}
                                                                                   {\footnotesize \textsuperscript{\textdagger} Trained from scratch without prior initialization from SIMPL marginal proposals. The SIMPL scene encoding is still used as context. \par}
                                                                                   \end{table*}

% ---------- Table 2: accelerated diffusion steps + runtime ----------
\begin{table*}[!t]
\caption{Accelerated DDIM sampling ablation with a fixed model and varying reverse diffusion steps. Runtime per \replanning\ step.}
\label{tab:accel_steps}
\centering
\small
\setlength{\tabcolsep}{4pt}
\renewcommand{\arraystretch}{1.0}
\resizebox{0.87\textwidth}{!}{%
\begin{tabular}{l c c c c c c}
\hline
\textbf{Method} & \textbf{Steps} & \textbf{Runtime [ms]} $\downarrow$ & \textbf{Collision [\%]} $\downarrow$ & \textbf{Off-road [\%]} $\downarrow$ & \textbf{Realism} $\downarrow$ & \textbf{ADE [m]} $\downarrow$\\
\hline
Ours (w/o Guidance) & 5  & \rtOursSfive       & \mCollOursSfive       & \mOffOursSfive       & \mRealOursSfive       & \mMinADEOursSfive\\
                                                                                    & 10 & \rtOursSten        & \mCollOursSten        & \mOffOursSten        & \mRealOursSten        & \mMinADEOursSten\\
                                                                                     & 25 & \rtOursStwentyfive & \mCollOursStwentyfive & \mOffOursStwentyfive & \mRealOursStwentyfive & \mMinADEOursStwentyfive\\
                                                                                     \hline
                                                                                     Ours (w/ Guidance) & 5  & \rtOursGuidedSfive       & \mCollOursGuidedSfive       & \mOffOursGuidedSfive       & \mRealOursGuidedSfive       & \mMinADEOursGuidedSfive\\
                                                                                      & 10 & \rtOursGuidedSten        & \mCollOursGuidedSten        & \mOffOursGuidedSten        & \mRealOursGuidedSten        & \mMinADEOursGuidedSten\\
                                                                                       & 25 & \rtOursGuidedStwentyfive & \mCollOursGuidedStwentyfive & \mOffOursGuidedStwentyfive & \mRealOursGuidedStwentyfive & \mMinADEOursGuidedStwentyfive\\
                                                                                       \hline
Ours (w/o Prior Init) & 5  & \rtNoPriorSfive       & \mCollNoPriorSfive       & \mOffNoPriorSfive       & \mRealNoPriorSfive       & \mMinADENoPriorSfive\\
 & 10 & \rtNoPriorSten        & \mCollNoPriorSten        & \mOffNoPriorSten        & \mRealNoPriorSten        & \mMinADENoPriorSten\\
 & 25 & \rtNoPriorStwentyfive & \mCollNoPriorStwentyfive & \mOffNoPriorStwentyfive & \mRealNoPriorStwentyfive & \mMinADENoPriorStwentyfive\\
\hline
\end{tabular}
}
\end{table*}
% --- Fig. 3 (qualitative results) ---
\begin{figure*}[!t]
  \centering
  \setlength{\abovecaptionskip}{1pt}
  \setlength{\belowcaptionskip}{0pt}

  % Light grey borders for all panels (consistent across rows)
  \setlength{\fboxsep}{0pt}
  \setlength{\fboxrule}{0.4pt}
  \newcommand{\panel}[1]{\fcolorbox{gray!60}{white}{#1}}

  \setlength{\tabcolsep}{2pt}
  \renewcommand{\arraystretch}{1}
  \newcommand{\vrowgap}{0.5pt}
  % Unified crop box and size for all 9 panels (force identical boxes)
  \newcommand{\panelimg}[1]{\panel{\includegraphics[width=4.55cm,height=4.55cm,trim=18 16 18 16,clip,keepaspectratio=false]{#1}}}
  % "Compressed" look for middle-row panels: render at lower effective resolution
  % Middle row uses a slightly smaller left trim so the timestamp box isn't cut off.
  \newcommand{\panelimgLow}[1]{\panel{\includegraphics[width=4.55cm,height=4.55cm,trim=6 12 12 12,clip,keepaspectratio=false]{#1}}}
  % Third row: match the exact box size, but use a gentler crop so the top-left time box stays visible.
  % Format: trim = left bottom right top (in bp).
  \newcommand{\panelimgCA}[1]{\panel{\includegraphics[width=4.55cm,height=4.55cm,trim=6 6 6 6,clip,keepaspectratio=false]{#1}}}
  \newcommand{\hpanelgap}{\hspace{4pt}}
  \begin{tabular}{cccc}
    \raisebox{0.65\height}{\rotatebox{90}{\small Nominal policy}} &
    \panelimg{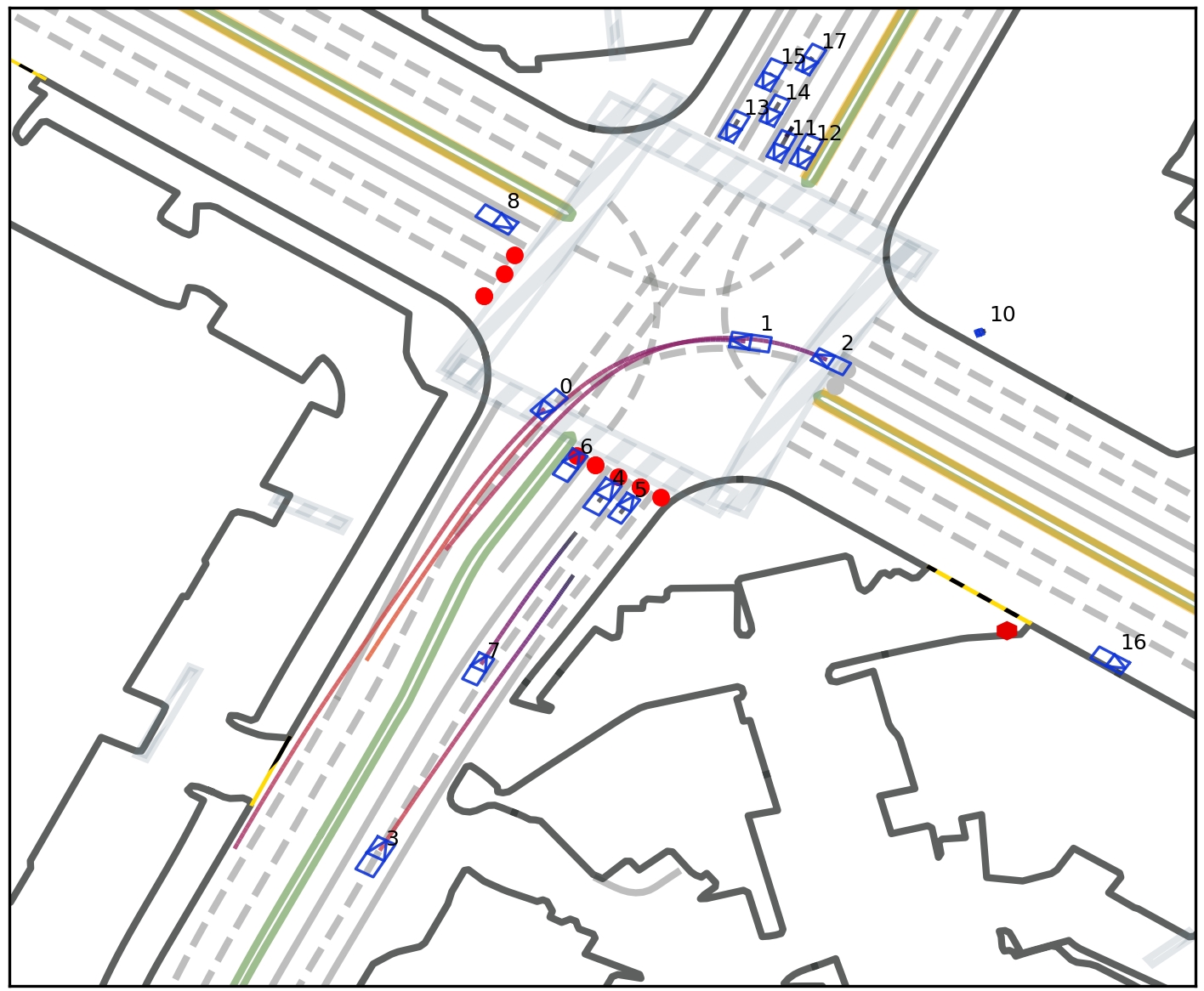}\hpanelgap &
    \panelimg{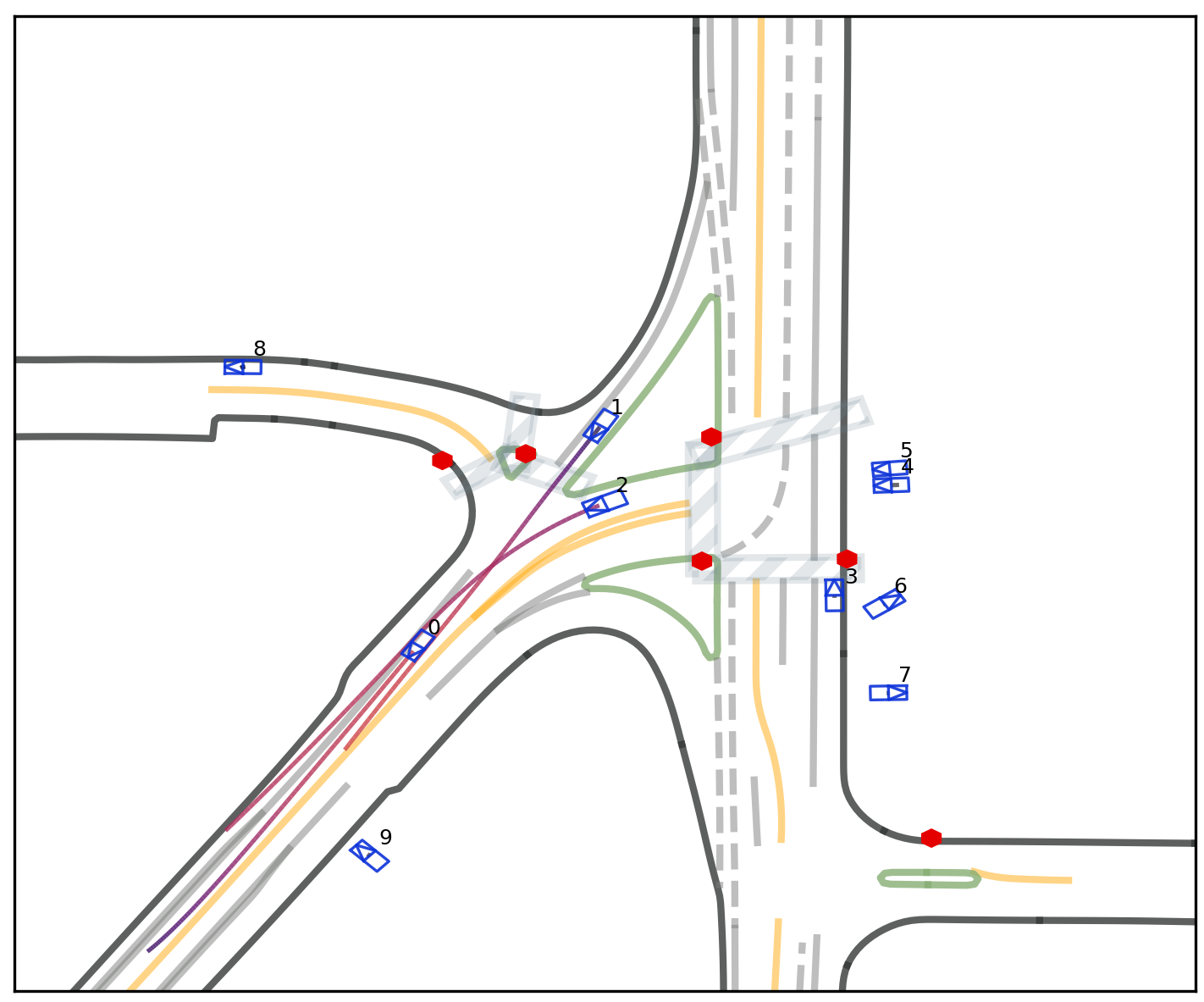}\hpanelgap &
    \panelimg{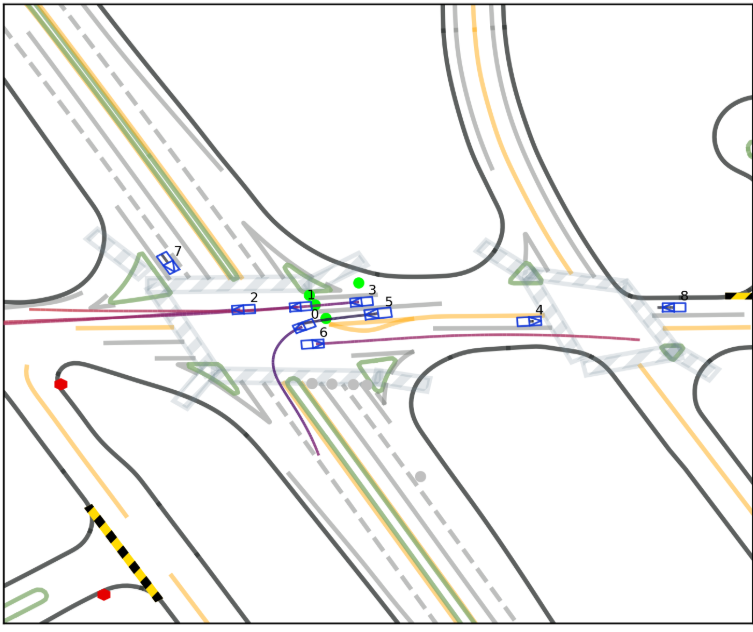} \\[\vrowgap]
    \raisebox{0.40\height}{\rotatebox{90}{\small Objective guidance}} &
    \panelimgCA{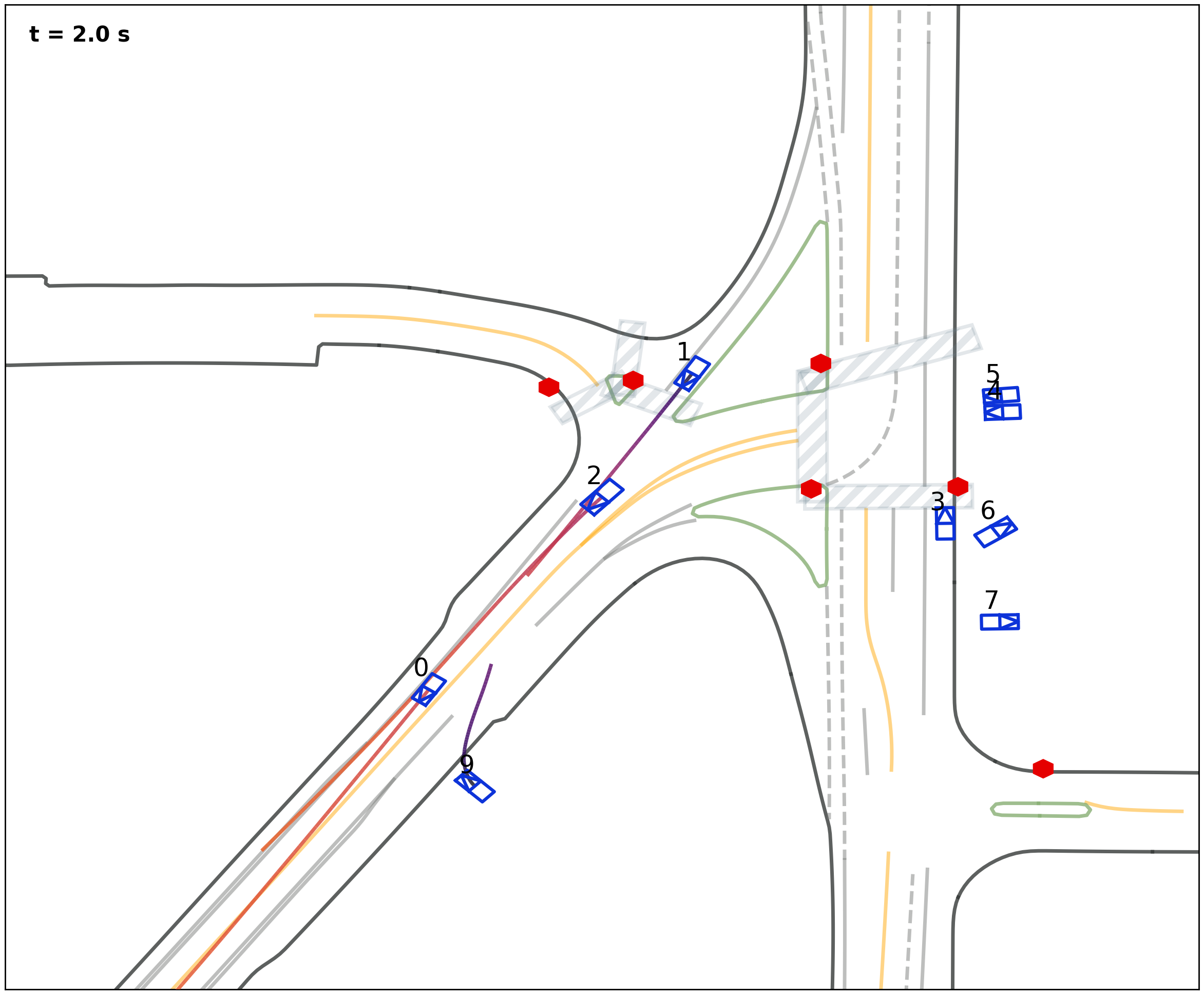}\hpanelgap &
    \panelimgCA{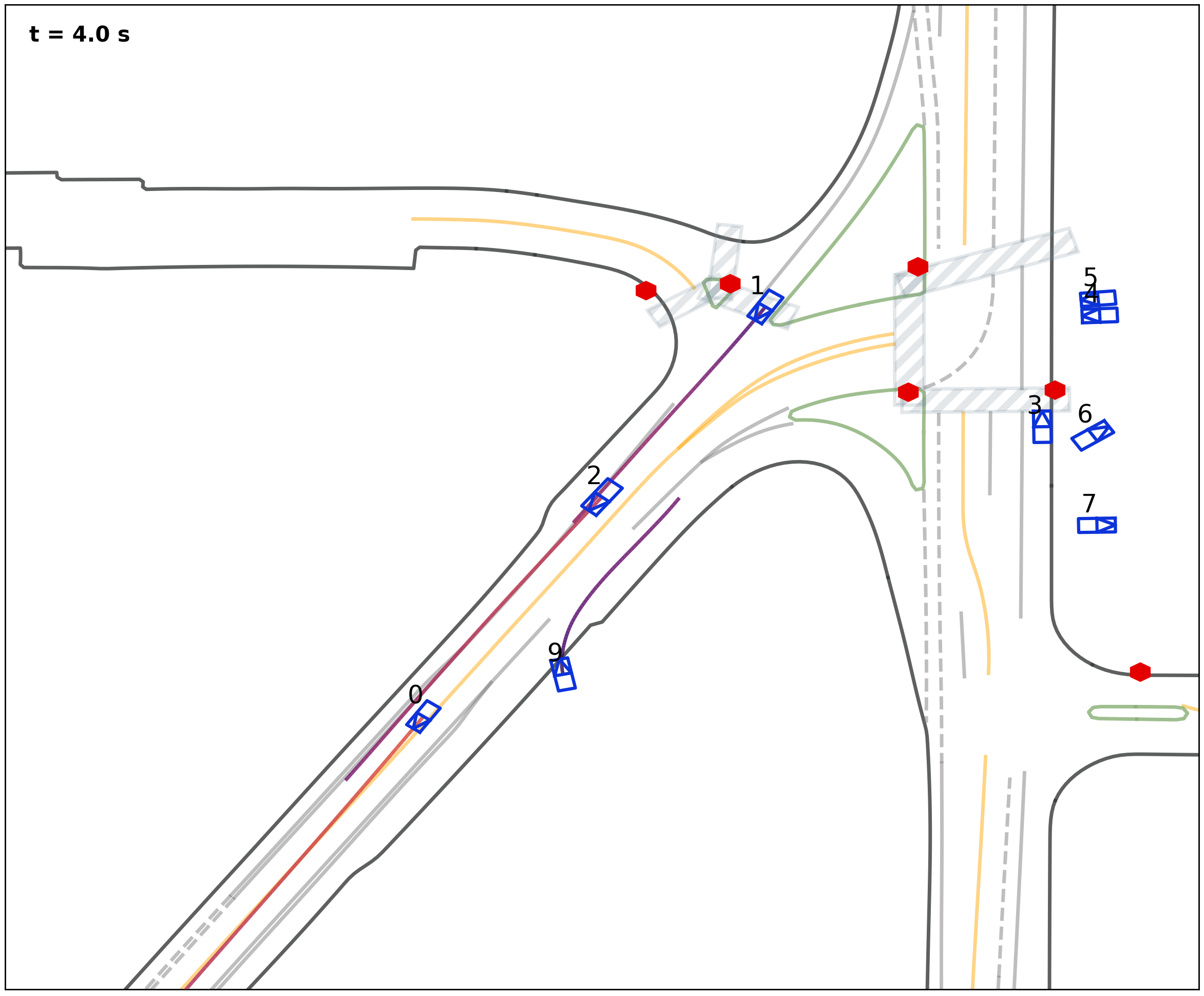}\hpanelgap &
    \panelimgCA{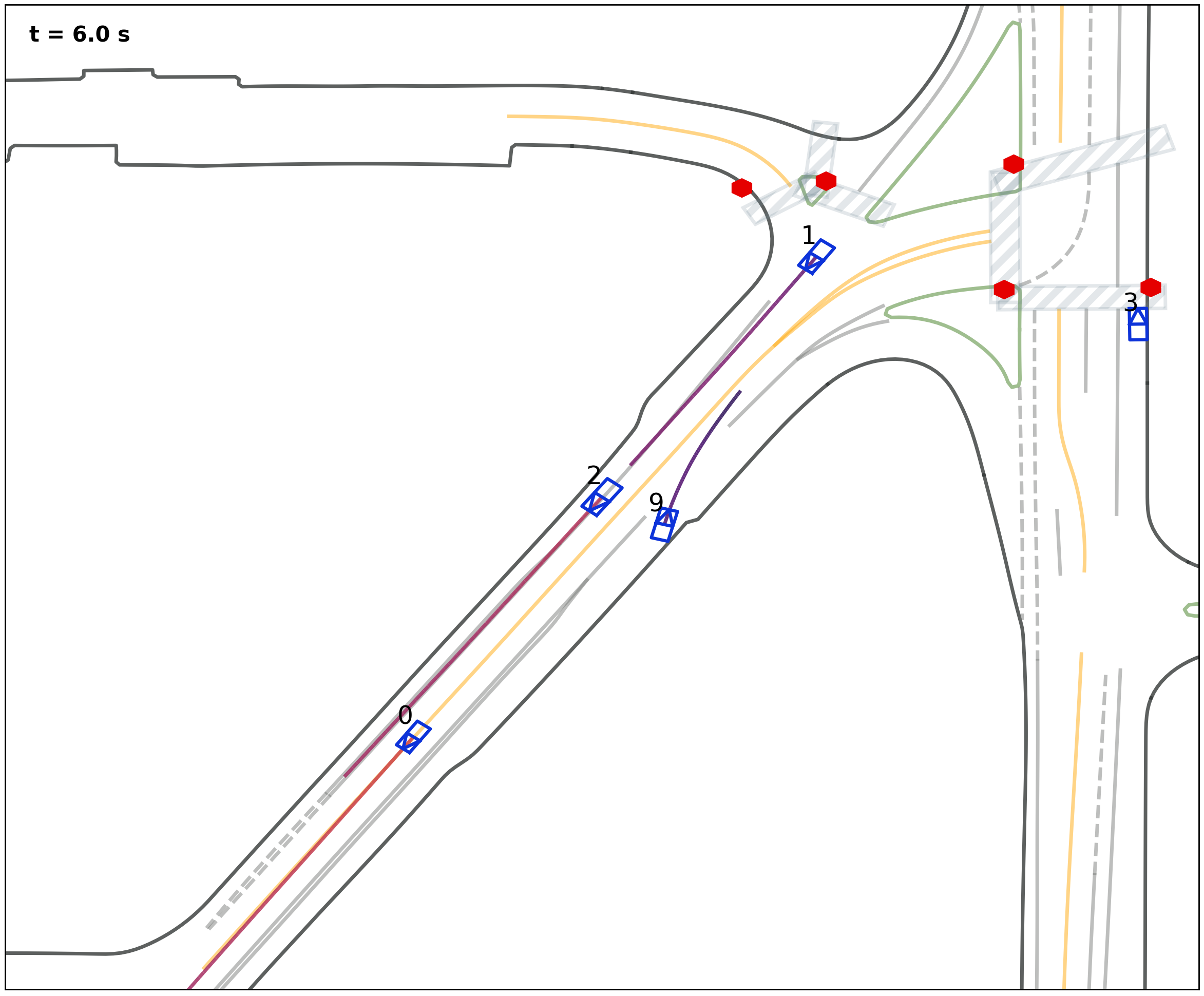} \\[\vrowgap]
    \raisebox{0.40\height}{\rotatebox{90}{\small Adversarial guidance}} &
    \panelimgCA{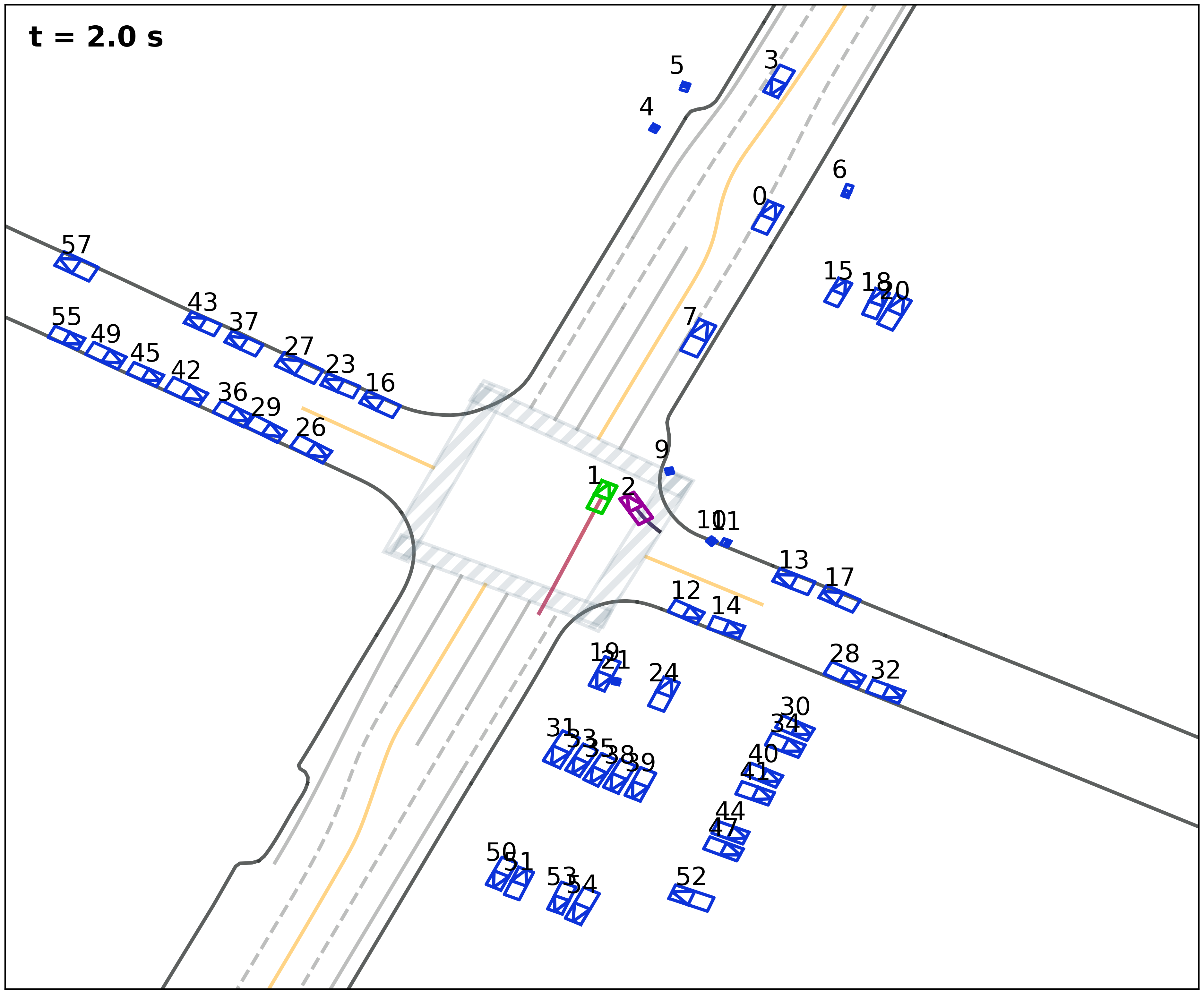}\hpanelgap &
    \panelimgCA{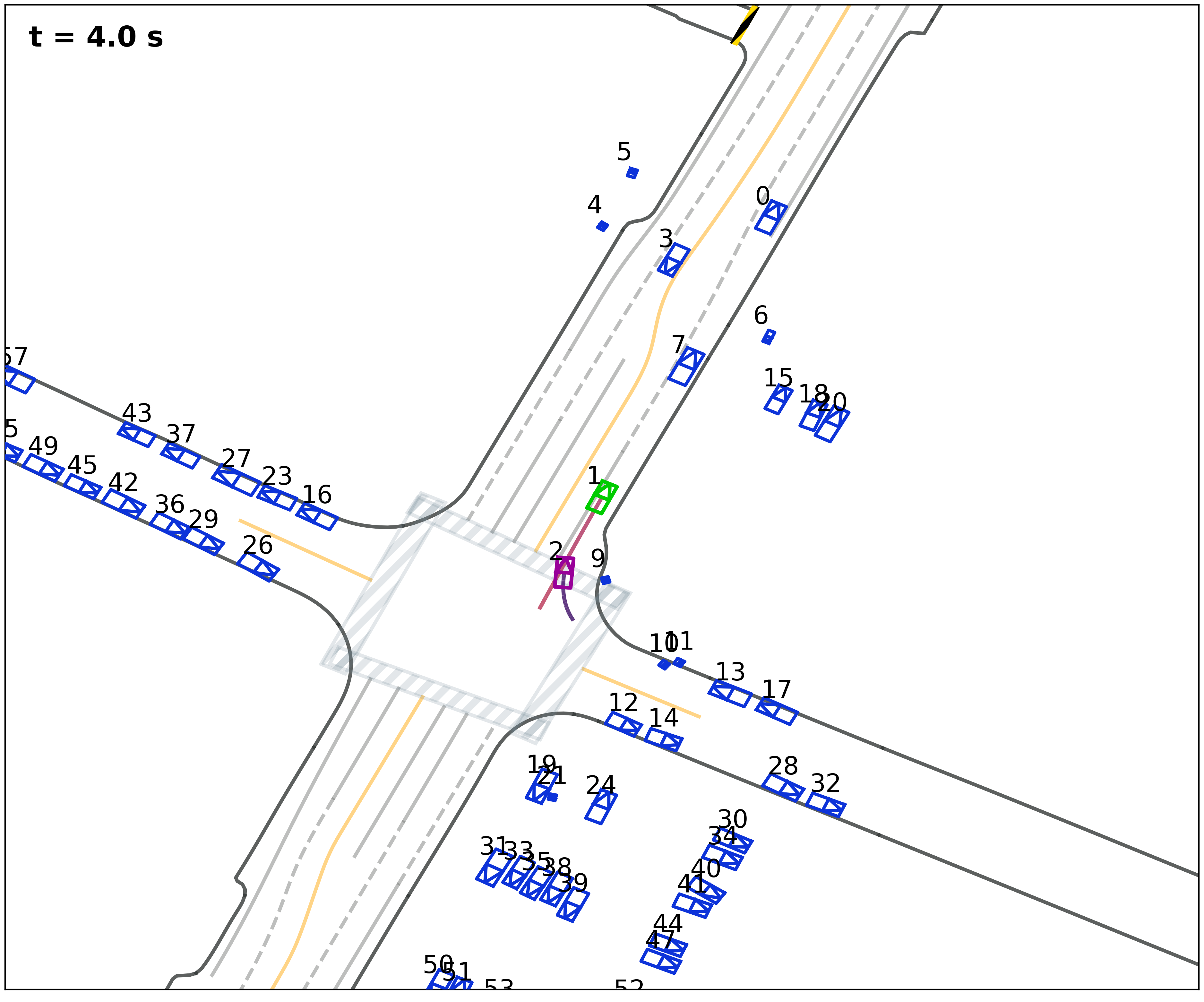}\hpanelgap &
    \panelimgCA{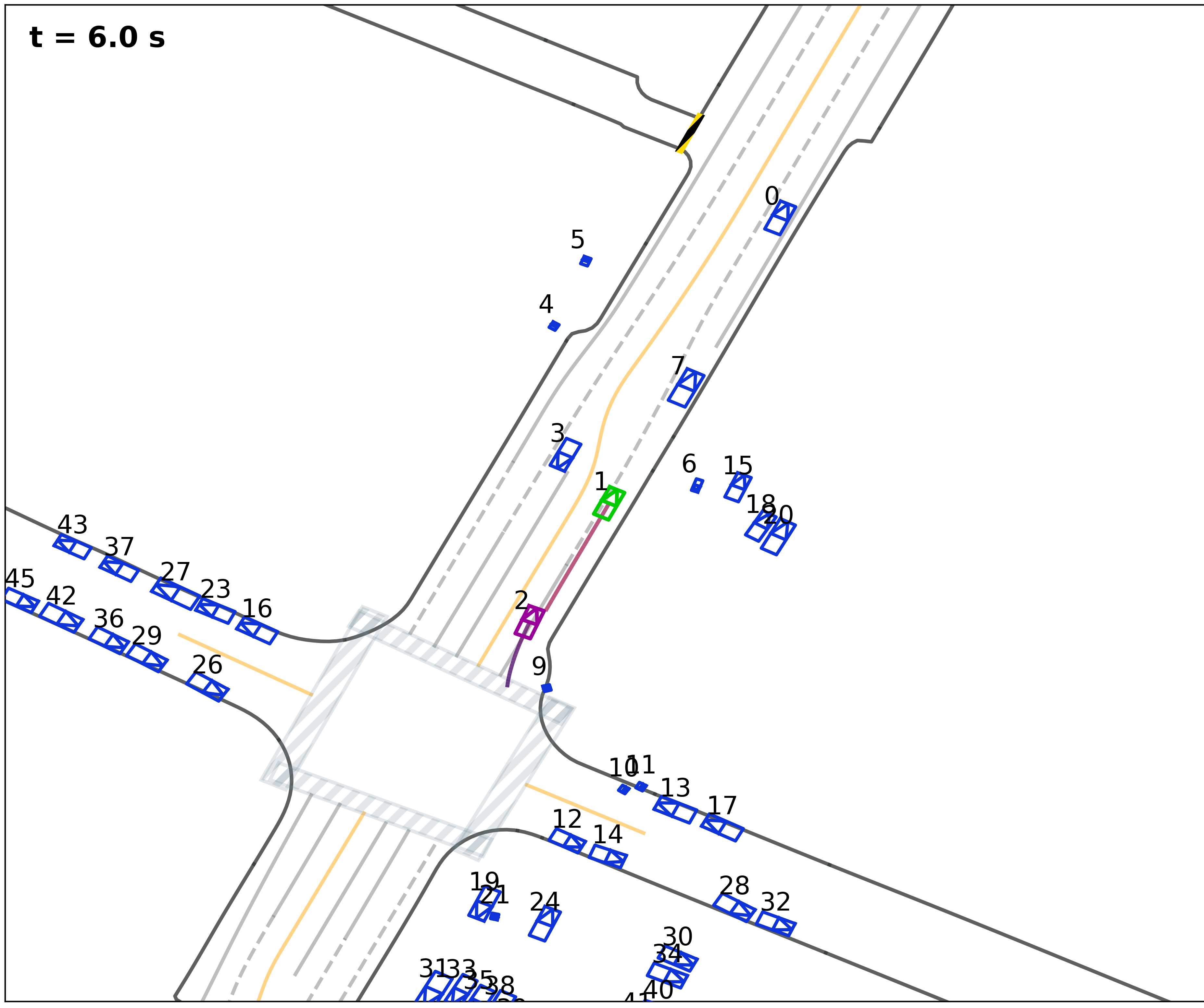} \\
  \end{tabular}

  \caption{Qualitative results. \textbf{Top:} Nominal policy (w/o guidance) our method produces scene-consistent predictions for many reactive agents that remain interactive and map-adherent. \textbf{Middle:} Objective-based guidance encourages stronger lane adherence and safer behaviors (agent~9 accelerating while other agents stay on the road). \textbf{Bottom:} Game-theoretic guidance where agent~2 (magenta) attacks the ego evader (green).}
  \label{fig:first_results}
\end{figure*}

% --- Safety-game guidance aggressiveness ablation table (separate) ---
\begin{table*}[!t]
  \caption{Safety-game guidance aggressiveness ablation (100 scenarios). We vary $w_{\mathrm{pursue}}$ and $\eta$ (fixed model and scenarios) and report mean $\pm$ std over $R=3$ seeds.}
  \label{tab:safety_aggressiveness}
  \centering
  \footnotesize
  \setlength{\tabcolsep}{3.5pt}
  \renewcommand{\arraystretch}{1.05}
  \resizebox{0.87\textwidth}{!}{%
\begin{tabular}{rcccccc}
  \hline
  $w_{\mathrm{pursue}}$ & $\eta$ & Ego Adv Coll [\%] $\uparrow$ & Adv Off-road [\%] $\downarrow$ & Realism $\downarrow$ & $\min_t d_{\mathrm{ego,adv}}$ [m] $\downarrow$ & Impact speed [m/s] $\uparrow$ \\
  \hline
  10 & 0.05 & $18.9 \pm 2.0$ & $2.8 \pm 0.6$ & $0.53 \pm 0.03$ & $2.78 \pm 0.10$ & $4.21 \pm 0.58$ \\
  30 & 0.07 & $33.6 \pm 2.6$ & $3.9 \pm 0.7$ & $0.58 \pm 0.03$ & $2.29 \pm 0.09$ & $6.05 \pm 0.67$ \\
  60 & 0.10 & $47.8 \pm 3.4$ & $5.8 \pm 0.9$ & $0.66 \pm 0.04$ & $1.91 \pm 0.08$ & $7.82 \pm 0.74$ \\
  \hline
  \end{tabular}
}
\end{table*}

                                                                                                                                                                                                                                               % Restore normal IEEE column balancing/vertical spacing for final pages.
                                                                                                                                                                                                                                               \flushbottom
                                                                                                                                                                                                                                               \normalfont\normalsize\normalcolor

                                                                                                                                                                                                                                               % Discussion continues immediately.
\begingroup
\input{discussion}
\endgroup

\begingroup
\input{conclusion}
\endgroup

\begingroup
\sloppy
\emergencystretch=1em
\FloatBarrier
\bibliographystyle{IEEEtran}
\bibliography{references}
\endgroup

\end{document}

%% file: related_work.tex
\section{Related Work}\label{sec:related}
This section reviews prior work on adversarial scenario generation, traffic simulation and diffusion-based motion modeling and highlights gaps that motivate the proposed approach.

\subsection{Adversarial Scenario Generation}
Safety-critical scenarios are infrequent in logged driving data, motivating approaches based on importance sampling~\cite{jiang2023efficient,most2024multi} and reinforcement learning~\cite{cao2022trustworthy,dagdanov2022defix} to identify failures. STRIVE~\cite{rempe2022generatingusefulaccidentpronedriving} introduced gradient-based adversarial optimization using a learned traffic prior and subsequent methods improved physical plausibility through kinematic constraints in KING~\cite{hanselmann2022kinggeneratingsafetycriticaldriving} and closed-loop adversarial training in CAT~\cite{zhang2023catclosedloopadversarialtraining}. Recent work has also leveraged human-driving priors for adversarial scenario generation~\cite{Hao_2024}. A recurring tension remains that optimizing exclusively for adversarial objectives can produce unrealistic behaviors whereas strict realism constraints may reduce adversarial effectiveness. Our guidance design addresses this tension by enabling safety edits while maintaining realism in closed-loop rollouts.
\subsection{Data-Driven Traffic Simulation}
Traffic simulation has moved from rule-based models such as IDM~\cite{Treiber_2000} and MOBIL~\cite{kesting1999general} toward data-driven policies and generative simulators trained on large-scale driving datasets.

Imitation-learning simulators that simultaneously roll out all agents such as TrafficSim~\cite{suo2021trafficsimlearningsimulaterealistic} and generative scenario augmentation methods such as TrafficGen~\cite{feng2023trafficgenlearninggeneratediverse} enhance realism and diversity. Other systems provide scalable evaluation simulators built on large datasets such as Waymax~\cite{gulino2023waymaxaccelerateddatadrivensimulator}. When paired with simple hand-designed policies (e.g., IDM), such platforms are better viewed as scalable evaluation backends than strong realism baselines. They also offer limited mechanisms for targeted safety-critical editing while preserving interactive closed-loop consistency. Our method aims to add this editing capability while keeping joint interaction fidelity under replanning.

\subsection{Diffusion Models for Scenario Generation}
Diffusion models are effective for multimodal trajectory generation and flexible conditioning. MotionDiffuser~\cite{jiang2023motiondiffusercontrollablemultiagentmotion} uses compressed representations to enable controllable joint prediction and variants such as MID~\cite{gu2022stochastictrajectorypredictionmotion} model trajectory ambiguity with Transformer denoisers. Intention-aware denoising has also been explored for trajectory prediction~\cite{liu2024intentionawaredenoisingdiffusionmodel}. To reduce inference cost, leapfrog diffusion initialization such as LED~\cite{mao2023leapfrogdiffusionmodelstochastic} aims to skip denoising steps. A practical limitation remains the computational demand at test time. Joint multi-agent diffusion can require many reverse steps and guidance that differentiates through the reverse process can further increase runtime. We address this by proposal-conditioned initialization and compact action latents that reduce steps without changing the model class.
Unlike~\cite{huang2024vbd} that refines trajectories starting from standard diffusion noise, our primary efficiency gain comes from proposal-conditioned initialization in a compact action latent, which changes the starting distribution and makes few-step DDIM viable in closed-loop replanning.

\subsection{Guided and Controllable Diffusion}
Controllable diffusion is commonly achieved via classifier guidance~\cite{dhariwal2021diffusionmodelsbeatgans} or classifier-free guidance~\cite{ho2022classifierfreediffusionguidance}. Composable diffusion~\cite{liu2023compositionalvisualgenerationcomposable} shows that multiple objectives can be combined at inference by summing guidance terms. In planning, Diffusion-ES~\cite{yang2024diffusionesgradientfreeplanningdiffusion} explores gradient-free optimization. Game-aware approaches explicitly model strategic interaction in GameFormer~\cite{huang2023gameformer} and motivate game-theoretic guidance for safety-critical scenario synthesis. Our formulation adapts these ideas to multi-agent closed-loop simulation with guidance applied in a compact latent space to control behavior without retraining.

Collectively, prior work highlights a recurring tension between realism, controllability and efficiency. We address it with proposal-conditioned joint diffusion that preserves interaction fidelity while enabling low-latency guidance for targeted safety edits.

%% file: discussion.tex
\vspace{2mm}
\vspace{2mm}
\section{Evaluation}
\label{sec:evaluation}

Tables~\ref{tab:metrics} and~\ref{tab:accel_steps} indicate a consistent pattern in closed-loop multi-agent simulation. Enforcing \emph{joint} scene consistency improves interactive feasibility compared to composing agents independently while map and kinematic fidelity remain sensitive to representation choices and training objectives.

\smallskip
\noindent\textbf{Closed-loop performance trends.}
The rule-based and marginal/autoregressive baselines provide reasonable rollouts but can exhibit interaction failures when multiple agents' futures are combined without an explicit joint consistency mechanism.
In contrast, joint diffusion denoising couples agents through shared refinement steps conditioned on the same scene context which tends to reduce multi-agent conflicts and improve interaction coherence in closed-loop rollouts. However, improvements are not uniform across metrics and residual map violations and kinematic artifacts remain.
For example, in Table~\ref{tab:metrics}, our \emph{unguided} model reduces the collision rate from 11.70\% to 4.83\% and reduces the off-road rate from 9.10\% to 3.27\% while maintaining comparable realism and positional accuracy. Qualitative rollouts are shown in \figref{fig:first_results}.
These findings suggest that interaction modeling is important but not sufficient on its own to guarantee strong map adherence and physically consistent motion. Performance also depends on accurate scene encoding, map priors and the rollout model assumptions.

Ablations without proposal-informed initialization further suggest that proposal priors help keep sampling stable by constraining the reverse process toward plausible regions of the trajectory space.
When this initialization is removed, the denoiser must rely more heavily on context-only conditioning which can increase sensitivity to sampling variance and exacerbate rare failure modes.

\smallskip
\noindent\textbf{Effects of accelerated sampling and step count.}
We evaluate sampling efficiency by varying the number of reverse DDIM denoising steps per replanning iteration, with all other model and scenario settings fixed and report closed-loop safety and fidelity metrics.
The DDIM ablation study in Table~\ref{tab:accel_steps} shows that modest denoising steps can preserve quality while reducing latency.
Runtime increases approximately in proportion to the number of denoising steps, whereas the benefit of additional refinement diminishes beyond a moderate step count.
With proposal-informed initialization, even a small number of reverse steps can preserve quality, whereas very few steps leave limited opportunity to resolve multi-agent conflicts and improve constraint satisfaction when initialization is removed.
At high denoising steps, additional computation may produce smaller improvements in average closed-loop metrics and may shift sampling behavior toward increased variability which does not necessarily improve mean collision or off-road rates.
In our experiments, 5 to 10 reverse denoising steps consistently provide the most favorable balance relative to larger step counts which add compute without proportional gains in closed-loop metrics.
Without proposal-informed initialization, the context-only variant generally needs more reverse steps to achieve comparable closed-loop quality, indicating slower convergence toward plausible joint trajectories.
This balance becomes increasingly important when guidance is enabled, as guidance increases computation per step and may dominate runtime at higher step counts.
Efficiency comes from denoising a compact joint action-latent, warm-starting the reverse process with proposal-informed initialization and using only a small number of denoising steps at test time which keeps replanning latency low without retraining.

\noindent\textbf{Guidance, Behavior and Limitations.}
Objective-based guidance provides a practical mechanism for targeted scenario editing and constraint tightening without retraining and it enables controllable behavior shifts while preserving interactive rollouts. In our experiments, it consistently improves safety-oriented outcomes when tuned with moderate weights. Its effectiveness depends on the differentiability and calibration of the chosen objectives.
Guidance can introduce compromises: stronger optimization pressure for a specific objective may slightly degrade other fidelity metrics if the objective is imperfectly aligned with the data distribution.
Moreover, gradient-based guidance introduces computational overhead and can be sensitive to hyperparameters such as step size and weighting.

\smallskip
\noindent\textbf{Guidance aggressiveness.}
Table~\ref{tab:safety_aggressiveness} varies the pursuit weight $w_{\mathrm{pursue}}$ and step size $\eta$ to control adversarial guidance strength.
More aggressive settings increase the ego and adversary collision rate and impact severity, including higher impact speed but also increase off-road rate and reduce realism. This balance is essential for testing a policy under different criticality levels.

%% file: conclusion.tex
\section{Conclusion}
\label{sec:conclusion}

This work presents a diffusion-based method for closed-loop traffic scenario generation that produces realistic, interactive multi-agent behavior while remaining controllable at test time. The method combines instance-centric scene conditioning with a joint latent diffusion policy to generate scene-consistent rollouts under replanning. A key component is proposal-informed Gaussian initialization. Rather than starting reverse diffusion from isotropic Gaussian noise, initialization occurs from a shifted Gaussian whose mean and scale are computed from a marginal proposal model. This biases sampling toward plausible behaviors and improves sampling quality for joint generation.

Closed-loop simulation results indicate that joint denoising improves interaction consistency relative to composing marginal rollouts while inference-time guidance enables targeted scenario editing including safety-critical and adversarial interactions. Overall, the framework supports challenging yet plausible scenario generation for evaluation. Future work includes reporting Waymo Open Sim Agents Challenge metrics using the official evaluator and submission protocol, exploring scene-level rollout losses that penalize multi-agent inconsistency and training with closed-loop policy learning. We also plan to incorporate language conditioning for more intuitive control over scenarios.